\documentclass[letterpaper, 10pt, conference]{ieeeconf}  % Comment this line out
\IEEEoverridecommandlockouts                              % This command is only
\usepackage{graphicx}
\usepackage{subfigure}
\usepackage{amssymb}
\usepackage{amstext}
\usepackage{amsmath}

\title{\LARGE \bf MPG - A Framework for Reasoning on 6 DOF Pose Uncertainty}

\author{
Wendelin Feiten \and Muriel Lang % <-this % stops a space
\thanks{Wendelin Feiten is with Corporate Technology, 
Intelligent Systems \& Control, Siemens AG, D-80200 Munich, Germany, 
Muriel Lang is with Ludwig-Maximilians-University, Munich, Germany}
}

\begin{document}
\maketitle
\thispagestyle{empty}
\pagestyle{empty}
%%%%%%%%%%%%%%%%%%%%%%%%%%%%%%%%%%%%%%%%%%%%%%%%%%%%%%%%%%%%%%%%%%%%%%%%%%%%%%%%
\begin{abstract}
Reasoning about the pose, i.e. position and orientation of objects is one of the 
cornerstones of robotic manipulation under uncertainty. In a number of joint research 
projects our group is developing a robotic perception system that perceives and models 
an unprepared kitchen scenario with many objects. Since no single sensor or measurement 
provides sufficient information, a technique is needed to fuse a number of uncertain
estimates of the pose, i.e. estimates with a widely stretched probability density 
function ($pdf$). \\
The most frequently used approaches to describe the $pdfs$ are sample based description and 
multivariate normal (Gaussian) distributions. Sample based descriptions in 6D 
can describe basically any type of $pdfs$, but they require a large number of samples and
there are no analytic formulae to fuse several $pdfs$. For Gaussian distributions these formulae
exist, but the Gaussian distributions are unimodal and don't model widely spread distributions well. \\
In this paper we present a framework for probabilistic modeling of
6D poses that combines the expressive power of the 
sample based description with the conciseness and algorithmic power of the Gaussian models.
As parameterization of the 6D poses we select the dual quaternions, i.e. any pose is represented
by two quaternions. The orientation part of a pose is described by a unit quaternion. 
The translation part is described by a purely imaginary quaternion.  
A basic probability density function over the poses is constructed by selecting a tangent 
point on the 3D sphere representing unit quaternions and taking the Cartesian set product of 
the tangent space with the 3D space of translations. In this 6D Euclidean space a 6D Gaussian
distribution is defined. Projecting this Gaussian back to the unit sphere and renormalizing induces a 
distribution over 6D poses, called a Projected Gaussian.\\
A convex combination of Projected Gaussians is called a Mixture of Projected Gaussians (MPG). The set 
of MPG can approximate the probability density functions that arise in our application, is closed under the
operations mentioned above and allows for an efficient implementation.
\end{abstract}

%%%%%%%%%%%%%%%%%%%%%%%%%%%%%%%%%%%%%%%%%%%%%%%%%%%%%%%%%%%%%%%%%%%%%%%%%%%%%%%%
\section{Introduction} \label{section:intro} 

A framework for reasoning on the 6D pose should allow for treating a 6D pose and 
a rigid motion in the same way. This is important for the propagation of information, 
e.g. a pose information taken by camera. The pose of this camera is in itself uncertain w.r.t. the common reference frame for several measurements.\\
The pose representation should use as few parameters as possible. This reduces the 
required memory space, and if more than the minimal number of parameters is used,
it makes the renormalization of the parameters easier. 
The parameters of a composition of rigid motions should follow from 
the parameters of the single motions in a simple way. Also, the parameters should 
have no singularities or discontinuities.\\ 
The representation of $pdfs$ on the parameter space should interface with the standard 
representations, i.e. on the one hand it should be possible to find a correspondence 
between a representation from the new framework and a sample based description of a $pdf$, 
and on the other hand a unimodal parametric description (like a Gaussian) should be
included in the new framework. Further on, we demand that it is easy to calculate an estimate based on two
$pdfs$ from the framework that describe the same pose. Last, but not least, it should be
straightforward to calculate the $pdf$ of a composition of rigid motions from the $pdfs$ of 
the individual rigid motions. \\
Since each position and orientation w.r.t a given coordinate system
is the result of a translation and a rotation. \textit{Position} and
\textit{translation} can be and will be used synonymously in this
paper, as well as \textit{orientation} and \textit{rotation}. 
Also, \textit{pose} and \textit{rigid motion} are used synonymously.\\

In Section \ref{section:previouswork} we will recapitulate various
approaches to the parametrization of rigid motions and
corresponding probability density functions. None of them fulfills
all requirements listed above, but they provide ingredients to our
synthesis. In Section \ref{section:probabilisticOrientation} we will
present our approach to probability density functions over rigid
motions. The relation between sample based descriptions and the 
MPG framework is described in Section \ref{section:SampleBased}, and 
the convergence properties are investigated in Section 
\ref{section:convergence}. We describe the implementation and 
experimental results in Section \ref{section:Implementation}.
In Section \ref{section:conclusion} we will recollect the
presented system and indicate directions of future work.

%%%%%%%%%%%%%%%%%%%%%%%%%%%%%%%%%%%%%%%%%%%%%%%%%%%%%%%%%%%%%%%%%%%%%%%%%%%%%%%%
\section{Related work} 
\label{section:previouswork}

The Mixture of Projected Gaussians (MPG) was first presented by Feiten et al. \cite{Feiten2009} 
and then expanded by Muriel Lang in her Diploma Thesis \cite{Lang2011}. 
See there for additional references on previous work concerning the parameterization
of the rotation in 3D and the rigid motion. 

The representation of rigid motions and especially of orientation in three dimensions 
is a central issue in various disciplines of arts, science and engineering. 
Rotation matrix, Euler angles, Rodrigues vector and unit quaternions are 
the most popular representations of a rotation in three dimensions. Rotation 
matrices have many parameters, Euler angles are not invariant under transforms 
and have singularities and Rodrigues vectors do not allow for an easy composition 
algorithm. Stuelpnagel \cite{Stuelpnagel1964} points out that unit quaternions are a 
suitable representation of rotations on the hypersphere $S_3$ with few parameters, 
but does not provide probability distributions. Choe \cite{Choe2006} represents the 
probability distribution of rotations via a projected Gaussian on a tangent space. 
He only deals with concentrated distributions and does not take translations into 
account. Goddard and Abidi \cite{Goddard1997, Goddard1998} use dual quaternions for motion 
tracking. In their observations the correlation between rotation and translation 
is captured also. The probability distribution over the parameters of the state 
model is a unimodal normal distribution. If the initial estimate is sufficiently 
certain and if the information that shall be fused to the estimate is sufficiently 
well focused this is an appropriate model. As can be seen in \cite{Kavan2006} from 
Kavan et al. dual quaternions provide a closed form for the composition of rigid 
motions, similar to the transform matrix in homogeneous coordinates. 
Antone \cite{Antone2001} suggests to use the Bingham distribution in order to 
represent weak information even though he does not give a practical algorithm 
for fusion of information or propagation of uncertain information. By now it is 
known that propagated uncertain information only can be approximated by Bingham 
distributions. Further Love \cite{Love2007} states that the renormalization of the 
Bingham distribution is computationally expensive. Glover \cite{Glover2011} also works 
with a mixture of Bingham distributions and recommends to create a precomputed 
lookup table of approximations of the normalizing constant using standard floating 
point arithmetic. Mardia et al. \cite{Mardia2007} use a mixture of bivariate von Mises 
distributions. They fit the mixture model to a data set using the expectation 
maximization algorithm because this allows for modeling widely spread distributions. 
Translations are not treated by them. To propagate the covariance matrix of a random 
variable through a nonlinear function, the Jacobian matrix is used in general. 
Kraft et al. \cite{Kraft2003} use therefore an unscented Kalman Filter \cite{Julier1997}. 
This technique would have to be extended to the mixture distributions.\\

From the analysis of the previous work, we synthesize our approach as follows:
We use unit quaternions to represent rotations in 3D, and dual quaternions to
obtain a concise algebraic description of rigid motions and their composition.
The base element of a probability distribution over the rigid motions is a Gaussian
in the 6D tangent space, characterized by the tangent point to the unit quaternions and
the mean and the covariance of the distribution. Such a base element is called a
Projected Gaussian.
We use Mixtures of Projected Gaussians to reach the necessary expressive power of the
framework.

\section{Pose uncertainty by Mixtures of Projected Gaussian distributions}
\label{section:probabilisticOrientation}

We assume that the quaternion as such is sufficiently well known to
the reader. In order to clarify our notation, at first some basics
are restated.

\subsection{Rigid Motion and Dual Quaternions}
\label{subsection:Rotation and Quaternions}

Let $\mathbb{H}$ be the quaternions, i.e 
\begin{equation}
 \mathbb{H} = \{q| q=a+i b + j c + k d\,\&\,  a,b,c,d \in \mathbb{R} \} 
\end{equation}
where \(a\) is the real part of the quaternion, and
the vector \(v=(b,c,d)^\top \) is the imaginary part. The quaternions can 
be identified with \(\mathbb{R}^4\) via the coefficients, \\
 $q=a+i \cdot b + j \cdot c + k  \cdot d \sim [a,b,c,d] $. \\
The norm and the conjugate are denoted by  $\|q\|$ and $\overline{q}$. 
Further we denote the unit quaternions by $\mathbb{H}_1$ and the imaginary
quaternions by $\mathbb{H}_i$.

Analogously to the way that unit complex numbers \(z = \cos (\phi
)+i \sin (\phi ) = e^{i \phi }\) represent rotations in 2D via the
formula \(p_{\text{rot}}=z p\) for any point \(p\in \mathbb{C}\),
unit quaternions represent rotations in 3D. 

A point
$\left(u,v,w\right)^\top$ in 3D is represented as the purely
imaginary quaternion \(p=i \cdot u + j \cdot v + k \cdot w\); a rotation around the
unit 3D axis \(v\) by the rotation angle $\theta $ is represented by the
quaternion \(q=\cos (\theta /2)+\sin (\theta /2) \left(i \cdot v_1 + j \cdot v_2 + k \cdot
v_3\right)\). 

The rotated point is obtained as
$p_{\text{rot}}=q*p*\overline{q}$ . Clearly, $q$ and $-q$ represent the
same rotation, so the set $U$ of unit quaternions is a double
coverage of the special orthogonal group $\text{SO}(3)$ of
rotations in 3D. The composition of rotations corresponds to the 
multiplication of the corresponding unit quaternions. 

The rigid motions in 3D can in a similar way be represented by dual quaternions.
Let's again clarify the notation. 

The ring of the dual quaternions with the dual unit $\epsilon$, which has the properties 
$\epsilon \cdot 1 = 1 \cdot \epsilon = \epsilon$ and $\epsilon^2 = 0$, is defined as: 
\begin{equation}
 \mathbb{H}_D = \{dq\ |\ dq=q_1 + \epsilon\cdot q_2 \,\&\,  q_1, q_2 \in \mathbb{H}\}
\end{equation}

As for the quaternions, the addition and the scalar multiplication are component wise. 
The multiplication follows from the properties of the dual unit. 
The quaternion conjugate (there is also a dual conjugate and a total conjugate) is given by
$ \overline{q_1 + \epsilon \cdot q_2} := \overline{q_1} + \epsilon \cdot \overline{q_2} $.

The rigid motion consists of a rotation part, represented by a rotation unit quaternion $q_r$, and a
translation part, represented by a purely imaginary translation quaternion $q_t$. A point 
$ p = \left(x,y,z\right)^\top $ is 
embedded in $\mathbb{H}_D$ by 
\begin{equation}
p_d := [1,0,0,0]+\epsilon \cdot [0,x,y,z].
\end{equation}
With these definitions, the dual quaternion representing the rigid motion is defined by
\begin{equation}
dq := q_r + \epsilon \frac{1}{2} \cdot q_t*q_r
\label{ridigmotiondq}
\end{equation}
A point transformed by a rigid motion is represented by 
\begin{equation}
dq**\,p_d**\,\overline{dq} 
\end{equation}
As before, a composition of rigid motions is represented by
the product of the corresponding dual quaternions. \\
Note that $\mathbb{H}_1 \times \mathbb{H}_i$ 
is a double coverage of $SE(3)$. 

\subsection{Base Element}
A mixture distribution generally consists of a convex combination of base elements. In 
our case, base elements are $pdfs$ that are induced from a Gaussian distribution on $\mathbb{R}^6$. 
This space can be interpreted as a linearization of $SE(3)$ w.r.t. the dual quaternion representation 
at a rotation represented by a unit quaternion $q_0 = [a,b,c,d]$. We take the tangent space 
in $\mathbb{R}^4$ to the unit sphere $S_3$ at the point $(a,b,c,d)^\top$. 
We provide it with the basis that we derive from the canonical basis $\left\{e_1,e_2,e_3,e_4\right\}$ 
in $\mathbb{R}^4$ by applying the quaternion formulation for rotations in $\mathbb{R}^4$ 
with unit quaternions $q_l$ and $q_r$: 
\begin{eqnarray}
\mathrm{rot}(p) & = & q_l*p*\overline{q_r} \\
b_{i} & := & q_0*e_i*\overline{q_{id}} \quad \mathrm{for} \ i = 1 \ldots 4 
\end{eqnarray}
Note that $b_1 \sim q_0$. The vectors $\left\{b_2,b_3,b_4\right\}$ are an orthonormal basis of the 
tangent plane. The coefficients of elements of the tangent plane, together with the coefficients of the 
translation part, constitute the $6D$ tangent space $TS_{q_0}$ of the rigid motion. 
This tangent space is mapped to the rigid motions by a central projection for the
rotation part and an embedding for the translation part. 

Let $(u,v,w,x,y,z)$ be a point in this tangent space. 
Then 
\begin{equation} 
q_r := \frac{1}{\sqrt{1+u^2+v^2+w^2}} \cdot \left(b_1 + u \cdot b_2 + v \cdot b_3 + w \cdot b_4\right)
\end{equation}
corresponds to a unit quaternion representing rotation. The imaginary quaternion 
\begin{equation}
q_t := \left[0, x, y, z\right]
\end{equation}
represents the translation. Then the dual quaternion according to (\ref{ridigmotiondq})
represents the corresponding rigid motion. Since the projections depend on 
$q_0$ we denote this mapping by 
\begin{equation}
\Pi_{q_0}: TS_{q_0} \longrightarrow S_3 \times \mathbb{R}^3 \sim SE(3)
\end{equation}
We extend $\Pi_{q_0}$ to be two-valued by 
letting both $q_r$ and $-q_r$ be in the image. Thus all points on $S_3$ are
reached except for those orthogonal to $q_0$.

With this mapping we can define the base element. 

\textbf{Definition:} Let $q_0$ be an arbitrary unit quaternion.
Further, let $TS_{q_0}$ be the tangent space defined above, and 
let $\mathcal{N}(\mu ,\Sigma )$ be a Gaussian
distribution on $TS_{q_0}$, calling the corresponding
probability density function $p_T$. 
Then the function $p_{SE(3)}$ on $S_3 \times \mathbb{R}^3 \sim SE(3)$ given by 
\begin{eqnarray}
p_{SE(3)}(m) & := & \frac{1}{C}\cdot p_{TS}\left(\Pi_{q_0}^{-1}(m)\right) \\
C & := & \int_{S_3 \times \mathbb{R}^3}  p_{TS}\left(\Pi_{q_0}^{-1}(m)\right) \, \mathrm{d}m 
\end{eqnarray} 
is a $pdf$ on $SE(3)$. For rigid motions $m_0$ with a rotation 
quaternion orthogonal to $q_0$ we let $p_{SE(3)}(m_0) := 0$ - this is a smooth completion. 
We call this type of distribution a Projected Gaussian or $PG$ and denote it by 
\begin{equation}
p_{SE(3)} \sim \mathcal{N} \left(q_0, \mu ,\Sigma \right)
\end{equation}

\begin{figure}[h]
\centering
\subfigure[ ]{
\includegraphics[height=.4\columnwidth]{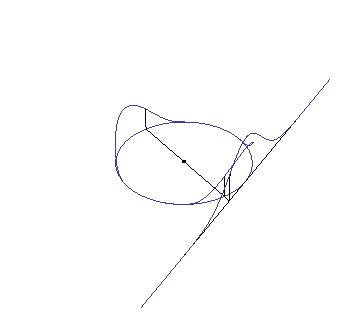}
}
\subfigure[ ]{
\includegraphics[height=.4\columnwidth]{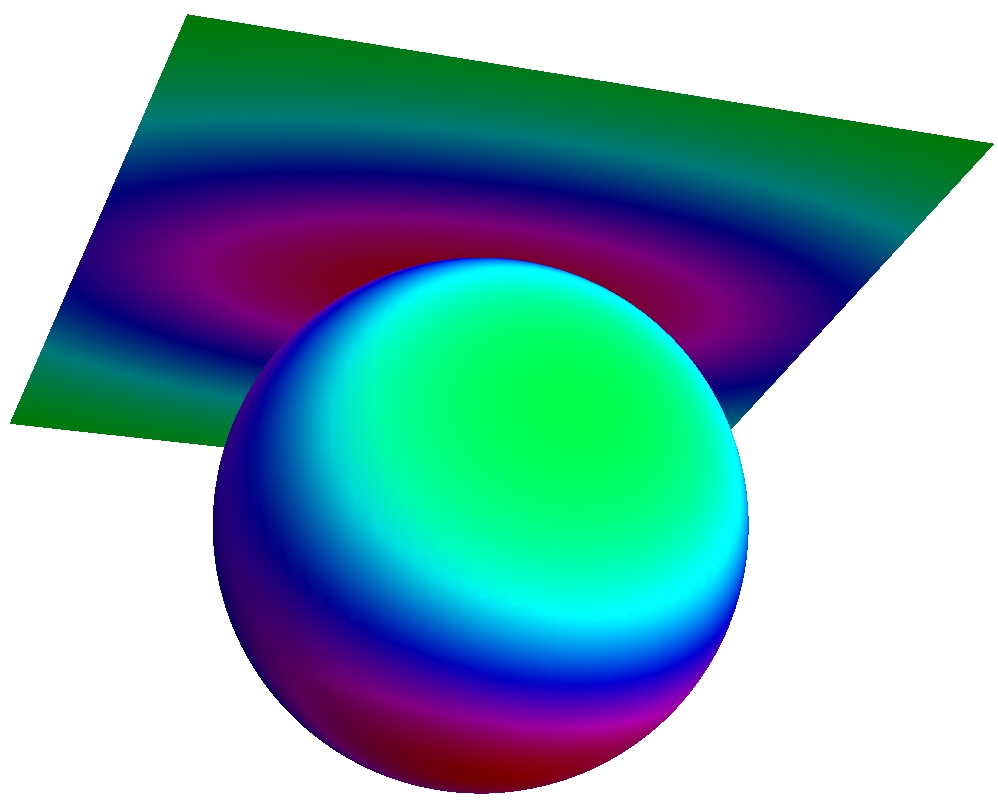}
}
\caption{(a) A base element on the unit circle, obtained by
projecting a Gaussian on a tangent line 
(b) An illustration of a Gaussian distribution projected to a 2D sphere}
\label{figure:projectedGaussian}
\end{figure}

See Figure \ref{figure:projectedGaussian} for 
examples of projected Gaussians. Note
that the peaks are lower due to renormalization.

The subset of $pdfs$ for which $\mu _u=\mu _v=\mu _w=0$ 
in the corresponding Gaussian on the tangent space is referred to
as $PG_0$.

Before extending our framework to Mixtures of Projected Gaussians, we 
would like to explain how to fuse and how to compose two Projected Gaussians.

\subsection{Fusion and Composition of Projected Gaussians}

Two Gaussian distributions $p_1\sim \mathcal{N}\left(\mu _1,\Sigma _1\right)$ 
and $p_2\sim \mathcal{N}\left(\mu _2,\Sigma_2\right)$ on $\mathbb{R}^n$ 
pertaining to the same phenomenon are fused by

\begin{equation}\label{usualfusion}
\begin{split}
p_3 & \sim \mathcal{N}\left(\mu_3,\Sigma_3\right) \\
\mu_3 & = \left(\Sigma _1 + \Sigma _2\right)^{-1} \cdot \left(\Sigma _1 \cdot \mu_2 + \Sigma_2 \cdot \mu_1 \right) \\
\Sigma_3 & = \left(\Sigma _1^{-1} + \Sigma _2^{-1}\right)^{-1} 
\end{split}
\end{equation} 

We can generalize this to $PGs$ 
$p_1 \sim \mathcal{N}\left(q_1, \mu_1, \Sigma_1 \right)$ and 
$p_2 \sim \mathcal{N}\left(q_2, \mu_2, \Sigma_2 \right)$
only if the tangents spaces are reasonably  
close - the angle between the normals to the tangent spaces should 
be less than 15${}^{\circ}$, or, equivalently by switching to the antipodal 
tangent point, larger than 165${}^{\circ}$. 

Then we define $q_3$ by renormalizing $q_1 + q_2$ to length 1 
and restate the original distributions approximately (we show
$p_1$, $p_2$ works the same way). With the Jacobian $J$ at the mean value $\mu_1$
of the mapping 
\begin{equation}
f = \Pi_{q_3}^{-1} \circ \Pi_{q_1} : TS_{q_1} \longrightarrow TS_{q_3}
\end{equation}
the parameters are estimated as 
\begin{equation}
\mu_{3,1} = f\left(\mu_1\right) \; \text{and} \; 
\Sigma_{3,1} = J \cdot \Sigma_1 \cdot J^\top
\end{equation}

Renormalization of the base elements involves an integral that is
hard to approximate with quadrature techniques. Therefore we use
Monte Carlo integration. Since the integrands are 
bounded by exponential functions that are easy to sample from, the 
integration is reasonably efficient. 
 
The estimates resulting from both original distributions are then fused on 
the common tangent space $TS_{q_3}$ using 
(\ref{usualfusion}). 
Generally, the rotation part of $\mu _3$ is not zero. Since it is advantageous to refrain 
to base elements of type $PG_0$,
$p_3$ is restated as above with the new tangent point $q_4=\Pi _{q_3}\left(\mu _3\right)$. 
The resulting probability density function on \(S_3\times \mathbb{R}^3\) needs to
be normalized according to Definition 1. 
The fused $pdf$ is denoted as $p_3 = \phi\left(p_1,p_2\right)$.

In robotics we frequently need to estimate the $pdf$ of a composition of 
two subsequent rigid motions given the $pdfs$ of the individual rigid motions 
(e.g. from an uncertain position in an uncertain camera frame to a position 
in world coordinates).  \\
Without loss of generality we refrain to $PG_0$ to define the composition, so
$p_1 \sim \mathcal{N}\left(q_1, 0, \Sigma_1 \right)$ and 
$p_2 \sim \mathcal{N}\left(q_2, 0, \Sigma_2 \right)$.

From the composition in terms of dual quaternions $dq_3=dq_2**\,dq_1$ the natural 
choice for the tangent point $q_3$ of the composition is $q_3 = q_2*q_1$. This
induces a mapping $g$ 
 
\begin{equation}
\begin{split}
g :&\, TS_{q_2} \times TS_{q_1} \longrightarrow TS_{q_3} \\
g\left(y_2,y_1\right) :&=\, \Pi_{q_3}^{-1} \left( \Pi_{q_2}\left(y_2\right)**\, \Pi_{q_1}\left(y_1\right)\right)
\end{split}
\end{equation}

Note that $g\left(0,0\right) = 0$, thus $\mu_3=0$. With

$\left.J_\gamma=\frac{\partial  g}{\partial
\left(y_2,y_1\right)}\right|_{(0,0)}$ and $\Sigma _\gamma=\left(
\begin{array}{cc}
 \Sigma _2 & 0 \\
 0 & \Sigma _1
\end{array}
\right)$ the resulting covariance matrix of the composition is
\(\Sigma _3=J_\gamma\cdot \Sigma _C\cdot J_\gamma^\top\).
The composition $p_3 \sim \mathcal{N}\left(q_3,0,\Sigma_3\right)$ is 
denoted by \\
$p_3 = \gamma\left(p_2,p_1\right)$.   

\subsection{Mixture of Projected Gaussians $MPG$}
\label{section:mpg}

As stated above, a precondition for the fusion of $PG$ base elements is
that their tangent points are sufficiently close to each other and
that they are sufficiently well concentrated. For this reason,
widely spread probability density functions should not be modeled in
a single base element.
Instead, we use a mixture of $PG$ base
elements. Thus let $p_i\in PG$ be
base elements, then a Mixture of Projected Gaussians is defined as 
\begin{equation}
p_m=\frac{1}{n}\sum_{i=1}^n \pi_i \cdot p_i \quad \text{with} \quad 0\leq \pi_i \leq 1 \quad \text{and} \quad \sum_{i=1}^n \pi_i=1
\end{equation}

Fusion and composition of $PGs$ carry over to $MPGS$ 
in a similar way as this work for Mixtures of
Gaussians \cite{Eidenberger2008}.

Let $p_{m,1}=\frac{1}{n}\sum _{i=1}^n \pi _{1,i} p_{1,i}$ and $p_{m,2}=\frac{1}{l}\sum _{j=1}^l \pi _{2,j} p_{2,j}$ be $MPGS$.
The fused mixture $p_{m,3}=\Phi\left(p_{m,1},p_{m,2}\right)$ is obtained by fusing and weighting the base elements of the original mixtures:
\begin{equation}
p_{m,3}= \frac{1}{C} \cdot \sum_{i,j=1}^{n,l} \lambda_{i,j}\cdot \pi_{1,i}\cdot \pi_{2,j}\cdot \phi\left(p_{1,i},p_{2,j}\right) \\ 
\end{equation} 
with a normalizing constant
\begin{equation}
C=\sum _{i,j=1}^{n,l} \lambda _{i,j}\cdot \pi _{1,i}\cdot \pi _{2,j}
\end{equation} 
The weights $\pi_{1,i}$ and $\pi_{2,j}$ are those of the
prior mixture. The plausibility is composed of two factors, \(\lambda _{i,j}=\alpha _{i,j}\cdot \delta _{i,j}\).

The factor $\alpha$ says whether the mixture elements can share a tangent space and thus probably pertain
to the same cases in the mixture (for detail see \cite{Lang2011}). The factor $\delta$ is the Mahalanobis distance of the mean values and covariances,  transported to the common tangent space.
\begin{equation*}
 \delta_{i,j} = \mathrm{e}^{-1/2\cdot \left(\mu _{3,1,i}-\mu _{3,2,j}\right)\cdot \left(\Sigma _{3,1,i}+\Sigma _{3,2,j}\right){}^{-1}\cdot \left(\mu _{3,1,i}-\mu _{3,2,j}\right)^\top}\\
\end{equation*} 
It expresses that even if the mixture elements could share a tangent space, they could still not be compatible.

The composition $p_{m,3} = \Gamma\left(p_{m,1},p_{m,2}\right)$ carries over 
in a similar manner. In this case, 
there is no question of whether two base elements could apply 
at the same time, since the two probability distributions are 
assumed to be independent, so the
factor  \(\lambda _{i,j}\) is omitted.

\begin{equation}
p_{m,3}=\frac{1}{C} \cdot \sum_{i,j=1}^{n,l}
\pi_{1,i}\cdot \pi_{2,j}\cdot \gamma\left(p_{1,i},p_{2,j}\right)
\end{equation}
with
\begin{equation} 
C=\left( \sum_{i,j=1}^{n,l} \pi_{1,i}\cdot \pi{2,j}\right).
\end{equation}

Note that in both cases the individually fused or combined resulting
base elements are assumed to be renormalized.

\section{$MPG$ and sample based description of $pdfs$}
\label{section:SampleBased}
The $MPGs$ try to fill the middle ground between sample based descriptions 
and unimodal parametric descriptions of $pdfs$. In our perception 
framework, we use sample based descriptions a lot, so we need to 
restate $pdfs$ available in one description also in the other one. 

Sampling from $MPG$ is easy, we first sample from the discrete $pdf$ induced 
by the weights, and then draw a sample for the Gaussian on the tangent space
using the Box-Muller algorithm. 

For fitting a $MPG$ to a sample set, we use a slight variant of the Expectation
Maximization Algorithm (see \cite{Bishop2007}):

\begin{enumerate}
\item Set the initial value for the means $\mu_i$, covariance matrices $\Sigma_i$ and weighting coefficients $\lambda_i$ and evaluate the log likelihood with these values.
\item E step:\\
Evaluate the responsibilities $\gamma(x_{n,i})$ using the current parameter values
$$\gamma(z_{j,i}) := \frac{\lambda_i\mathcal{N}(x_j|TS_i, \mu_i,\Sigma_i)}{\sum_{k}{}{\lambda_k\mathcal{N}(x_j|TS_k, \mu_k,\Sigma_k)}}$$
\item M step: \\
 Reestimate the parameters using the current responsibilities 
 $$\mu_i^{new} = \frac{1}{N_i} \sum_{j=1}^{N}{\gamma(z_{j,i})\cdot x_j}$$
 $$\Sigma_i^{new} = \frac{1}{N_i} \sum_{j=1}^{N}{\gamma(z_{j,i})(x_j-\mu_i^{new})(x_j-\mu_i^{new})^\top}$$ $$\lambda_i^{new} = \frac{N_i}{N}$$
 where $N_i = \sum_{j=1}^{N}{\gamma(z_{j,i})}$
\item Evaluate the log likelihood: 
$$\ln \mathrm{P}(X|TS, \mu, \Sigma, \lambda)\! =\!\! \sum_{j=1}^{N}{\ln\! \left(\!\sum_{i=1}^n{\lambda_i \mathcal{N}(x_j|TS_i, \mu_i,\Sigma_i)}\!\!\right)}$$
and check for convergence of either the parameters or the log likelihood. If the convergence criterion is not satisfied return to the E step.\\
\end{enumerate}
Note that we want to keep our base elements in $PG_0$, so re-estimating $\mu_i$ in the 
M-Step also means resetting the tangent points $q_i$. 

\section{Convergence Properties of the $MPG$}
\label{section:convergence}
The good news of the $MPG$ framework is that that we get good 
approximations with few base elements, and that it allows for fusing and 
composing $pdfs$ without having to go back to the original sensor readings.\\
The bad news is that the number of base elements tends to increase quadratically
with these operations. However, we can drop base elements with small weights 
and we can merge base elements with similar statistics without loosing much 
information, while decreasing the number of base elements.

In order to use the framework for the task of grasping, we defined
a grasp criterion: Given a $pdf$ $p$ of an object pose and a 
box $B \subset SE(3)$ that captures the
tolerances associated with the grasping task (e.g. width of the gripper, 
possible grasp on the object), we try to find a rigid transform $m$ that maximizes 
the probability of successfully grasping:
$$m:=\arg \max_{m \in SE(3)} \int_{m(B)} p(x) d\mu_{m(B)}$$

If $p_M(x)=\sum_{i=1}^n \lambda_i\cdot p(\mu_i,\Sigma_i,x)$, 
the integral is not very much affected
by dropping a base element with a small weight, let's assume the last one. 
Renormalizing the remaining weights
$$ \lambda _{i}^{'} := \frac{ \lambda _i }{1 - \lambda _n } $$
we get 
$$ \left|\int\limits_{B} \sum_{i=1}^n \lambda_i p_i \,
\mathrm{d}\mu_B - \int\limits_{B} 
\sum_{i=1}^{n-1}\lambda'_ip_i \,\mathrm{d}\mu_B \right| 
< 2\lambda_n$$
for any box $B$ (for a proof see \cite{Lang2011}).

The information loss due to combining two similar base elements is 
investigated in terms of a symmetric version of the Kullback-Leibler 
divergence, inspired by an investigation of Runnalls for 
ordinary mixtures of Gaussians (see \cite{Runnalls2007}). 
Let's assume the base elements 
$p_1 \sim \mathcal{N}\left(TS_{q_1},\mu_1,\Sigma_1\right)$ and
$p_2 \sim \mathcal{N}\left(TS_{q_2},\mu_2,\Sigma_2\right)$
from a $MPG$ $M$
have already been restated to the same tangent space, i.e. $q_1 = q_2$.
Then the symmetric Kullback-Leibler divergence between them is 
\begin{equation}
\begin{split}
\mathcal{D} =& (\Sigma_1^{-1}+\Sigma_2^{-1})(\mu_1-\mu_2)(\mu_1-\mu_2)^\top \\
d_{sKL}(p_1,p_2) =& \frac{1}{2}\mathrm{tr} \left( \Sigma_2^{-1}\Sigma_1+\Sigma_1^{-1}\Sigma_2 \quad + \mathcal{D} \right)-6
\end{split}
\end{equation}
If we now replace both $p_1$ and $p_2$ with the combined base element 
$$ p' = \frac{1}{\lambda_1 + \lambda_2} \cdot \left(\lambda_1 p_1 + \lambda_2 \cdot p_2 \right) $$
to obtain the modified $pdf$ $M'$ with less base elements, then we have:
\begin{equation}
d_{sKL}(M,M') \leq \frac{1}{\lambda_1 + \lambda_2}\cdot \left(\lambda_1 d_{sKL}(p_1,p') + \lambda_2 d_{sKL}(p',p_2) \right) 
\end{equation}
For details and proofs see \cite{Lang2011}.

\section{Implementation and Experimental Results}
\label{section:Implementation}
\begin{figure}[ht]
\centering
\subfigure[ ]{
\includegraphics[height=.4\columnwidth]{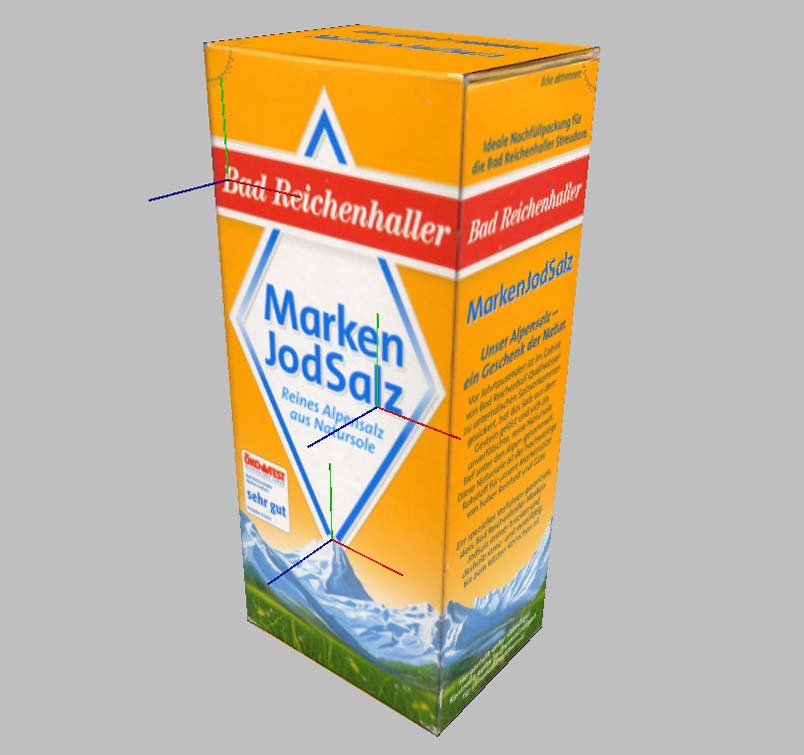}
}
\subfigure[ ]{
\includegraphics[height=.4\columnwidth]{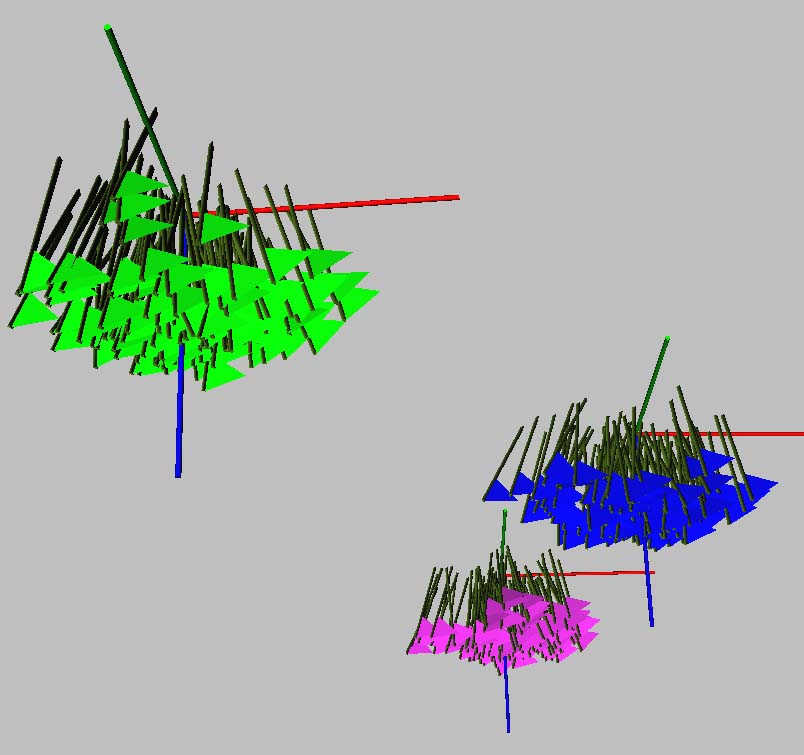}
}
\caption{(a) salt box (b) Stereo SIFT features on the salt box. 
The features have an orientation around the sight line, but the visibility range 
is assumed to be $15^\circ$, the translation invariance is low. Each feature 
is represented by a $MPG$ with seven base elements.}
\label{figure:SIFTfeature}
\end{figure}

The $MPG$ framework is fully implemented in Python. The probability density
functions are visualized by drawing random samples and for each pose 
painting a flag on the screen. The foot of the flag represents position, the
pole represents the z-axis of the rotated coordinate system, and the tip of 
the pennant marks the x-axis. 
As an application example we demonstrate how to estimate the pose of an object
based on SIFT features, e.g. the salt box of figure \ref{figure:SIFTfeature}. 
Let's assume that the robot detects the features 'B' of the word \textit{Bad} (green)
and 'l' of \textit{Salz} (blue). The mountain top will be used as a third feature (purple).\\

\begin{figure}[ht]
\centering
\subfigure[ ]{
\includegraphics[height=.4\columnwidth]{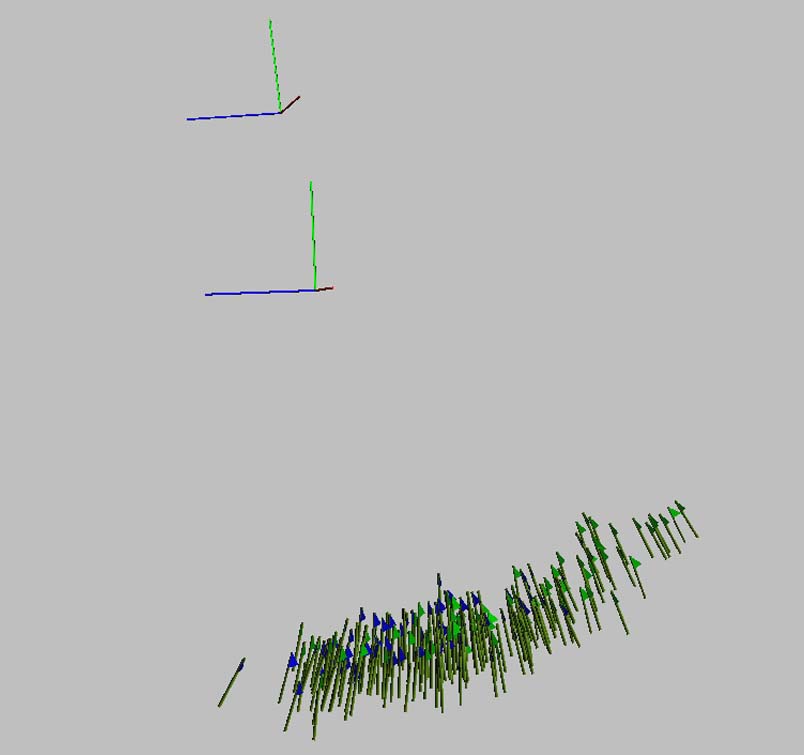}
}
\subfigure[ ]{
\includegraphics[height=.4\columnwidth]{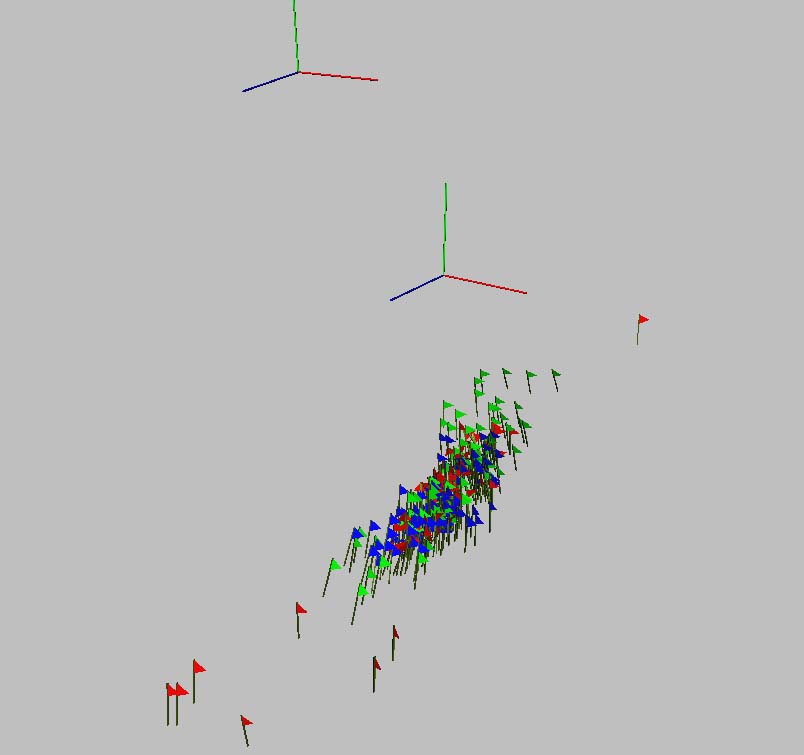}
}
\caption{(a) From the first two SIFT features, two distributions for the 
object pose are derived by the composition of the rigid motions camera to 
feature and feature to object coordinate system (b) The two distributions are fused (red).
 }
\label{figure:positionestimate2}
\end{figure}

\begin{figure}[ht]
\centering
\subfigure[ ]{
\includegraphics[height=.4\columnwidth]{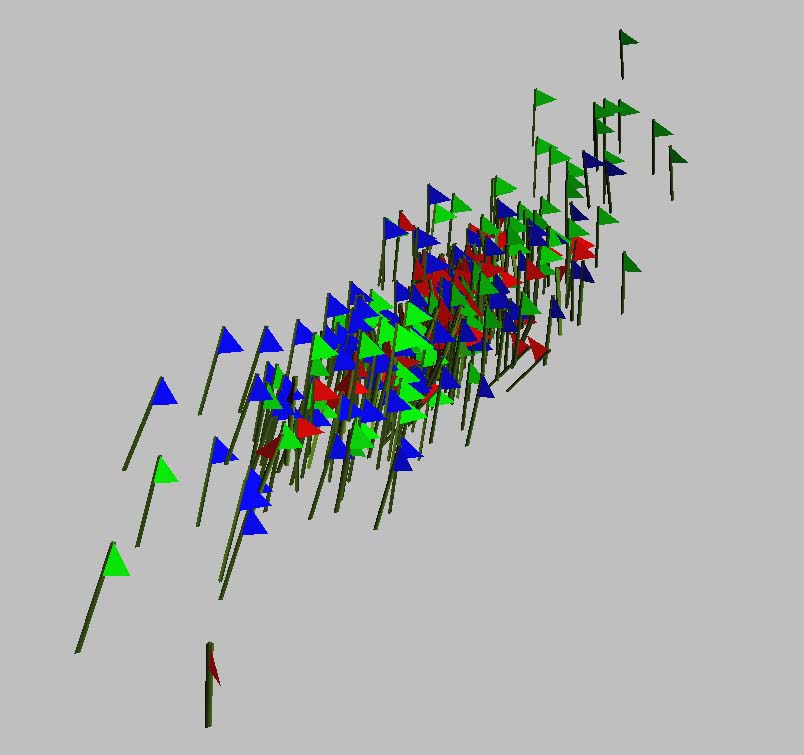}
}
\subfigure[ ]{
\includegraphics[height=.4\columnwidth]{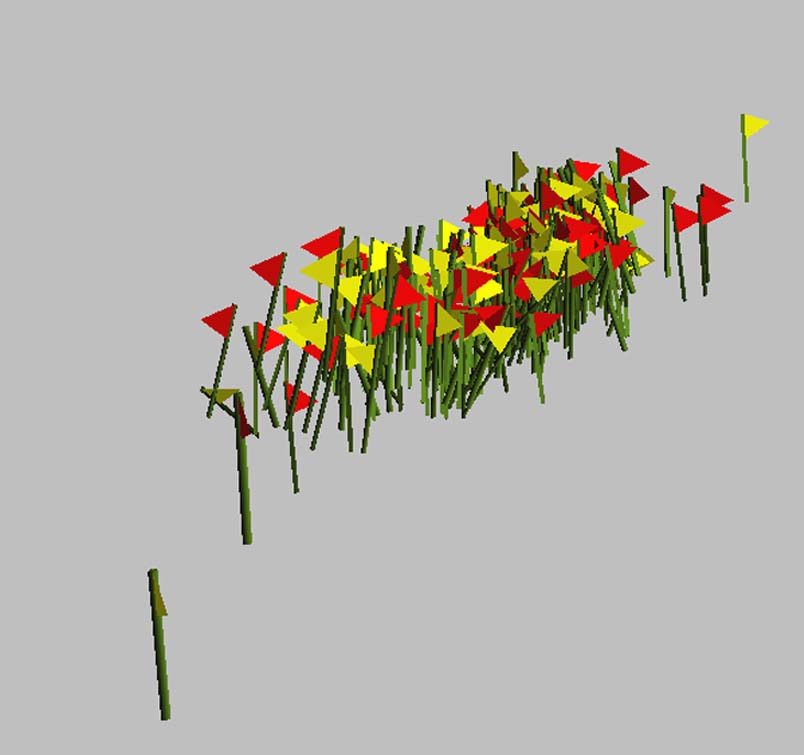}
}
\caption{(a) From originally 49 base elements, ten with low weights are skipped 
(b) The base elements are merged until only ten are left.
 }
\label{figure:positionskipped}
\end{figure}

\begin{figure}[ht]
\centering
\subfigure[ ]{
\includegraphics[height=.4\columnwidth]{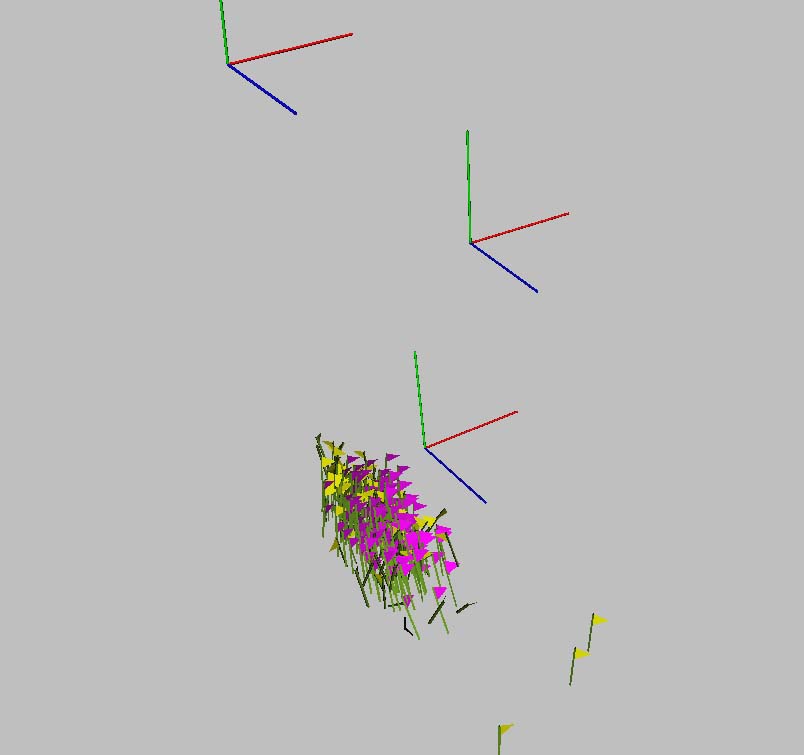}
}
\subfigure[ ]{
\includegraphics[height=.4\columnwidth]{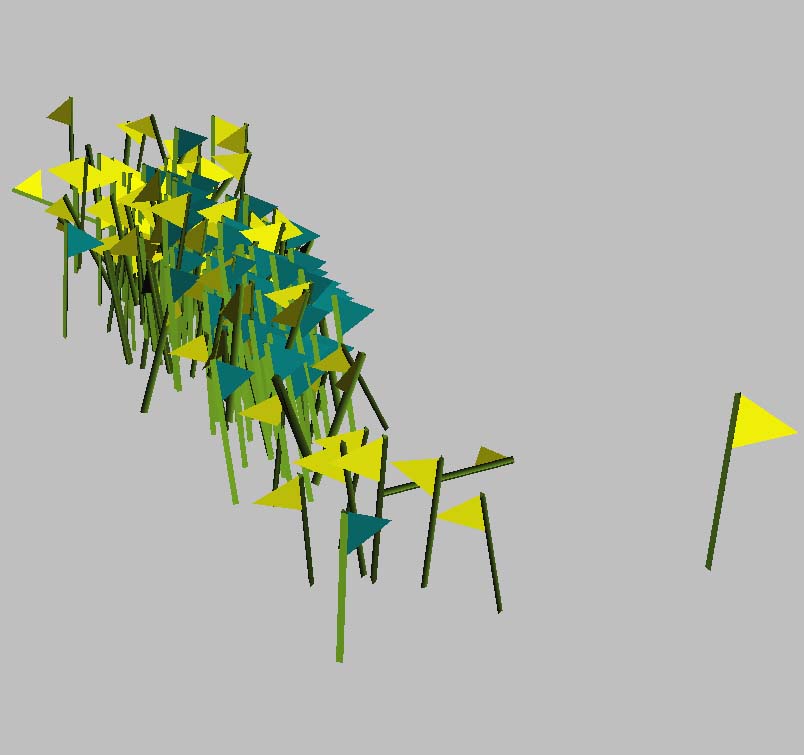}
}
\caption{(a) The object pose estimate from the next feature is added ... 
(b) ... and fused with the previous estimate (turquoise).
 }
\label{figure:measurement3}
\end{figure}

\begin{figure}[ht]
\centering
\subfigure[ ]{
\includegraphics[height=.4\columnwidth]{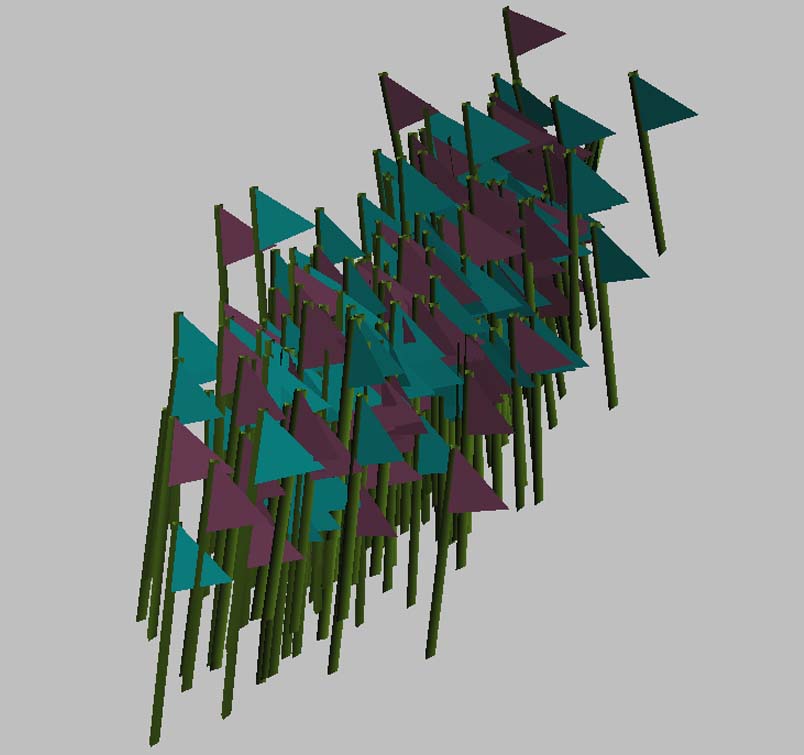}
}
\subfigure[ ]{
\includegraphics[height=.4\columnwidth]{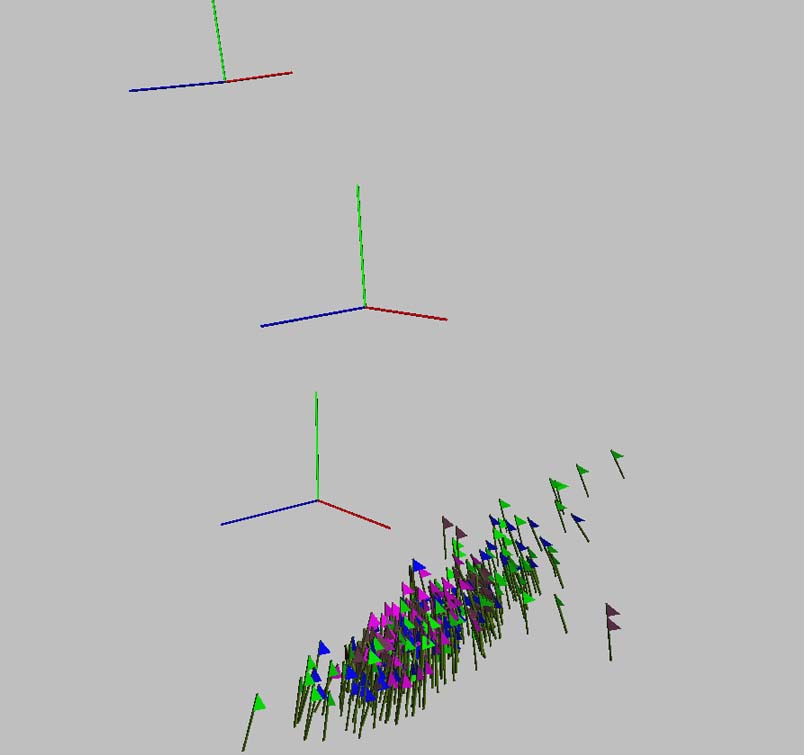}
}
\caption{(a) After merging again, the pose estimate converges. 
(b) Compare to the original estimates.
 }
\label{figure:final}
\end{figure}

\section{Conclusion and outlook}
\label{section:conclusion}

In this paper we present the framework of Mixtures of Projected
Gaussians that allows for modelling a large variety of possible
probability distribution functions of 6D poses. In contrast to a
sample based description, much fewer parameters are needed to
describe the distribution. The framework interfaces well with the 
sample based descriptions, and provides the inference mechanisms
of fusing and composing uncertain pose information.

The operations of fusion, propagation or multiplication of MPG distributions generally
result in a large number of mixture elements. However, many of them
have practically zero weight, while others are approximately identical. 
This is used to drop some base elements, and to merge others, thus reducing
their number without losing much information.

The algorithms for probabilistic inference (fusion, propagation, multiplication) are fully implemented in \textit{Python}.

The covariance matrices are currently estimated using the Jacobian of the non-linear transforms. 
These estimates could be improved by using the unscented
estimation technique (see Julier and Uhlmann \cite{Julier1997}).

In this paper we focus on the perception of static objects. The MPG
framework can be extended to the dynamic case as well, following
concepts by Goddard \cite{Goddard1998} and by Brox et al.  \cite{Brox2006}.

%%%%%%%%%%%%%%%%%%%%%%%%%%%%%%%%%%%%%%%%%%%%%%%%%%%%%%%%%%%%%%%%%%%%%%%%%%%%%%%%
\section{Acknowledgement}

This work was made possible by funding from the ARTEMIS Joint Undertaking as part of the project R3-COP and from the German Federal Ministry of Education and Research (BMBF) under grant no. 01IS10004E.

%%%%%%%%%%%%%%%%%%%%%%%%%%%%%%%%%%%%%%%%%%%%%%%%%%%%%%%%%%%%%%%%%%%%%%%%%%%%%%%%
\bibliography{referencesICRA}

% Generated by IEEEtran.bst, version: 1.14 (2015/08/26)
\begin{thebibliography}{10}
\providecommand{\url}[1]{#1}
\csname url@samestyle\endcsname
\providecommand{\newblock}{\relax}
\providecommand{\bibinfo}[2]{#2}
\providecommand{\BIBentrySTDinterwordspacing}{\spaceskip=0pt\relax}
\providecommand{\BIBentryALTinterwordstretchfactor}{4}
\providecommand{\BIBentryALTinterwordspacing}{\spaceskip=\fontdimen2\font plus
\BIBentryALTinterwordstretchfactor\fontdimen3\font minus
  \fontdimen4\font\relax}
\providecommand{\BIBforeignlanguage}[2]{{%
\expandafter\ifx\csname l@#1\endcsname\relax
\typeout{** WARNING: IEEEtran.bst: No hyphenation pattern has been}%
\typeout{** loaded for the language `#1'. Using the pattern for}%
\typeout{** the default language instead.}%
\else
\language=\csname l@#1\endcsname
\fi
#2}}
\providecommand{\BIBdecl}{\relax}
\BIBdecl

\bibitem{Feiten2009}
\BIBentryALTinterwordspacing
W.~Feiten, P.~Atwal, R.~Eidenberger, and T.~Grundmann, ``{6D Pose Uncertainty
  in Robotic Perception},'' in \emph{Advances in Robotics Research},
  T.~Kr{\"{o}}ger and F.~M. Wahl, Eds.\hskip 1em plus 0.5em minus 0.4em\relax
  Berlin, Heidelberg: Springer Berlin Heidelberg, 2009, pp. 89--98. [Online].
  Available: \url{http://link.springer.com/10.1007/978-3-642-01213-6{\_}9}
\BIBentrySTDinterwordspacing

\bibitem{Lang2011}
\BIBentryALTinterwordspacing
M.~Lang, ``{Approximation of Probability Density Functions on the Euclidean
  Group Parametrized by Dual Quaternions},'' arXiv preprint: arXiv:1707.00532,
  Ludwig-Maximilians-Universität München, 2011. [Online]. Available:
  \url{https://arxiv.org/abs/1707.00532}
\BIBentrySTDinterwordspacing

\bibitem{Stuelpnagel1964}
\BIBentryALTinterwordspacing
J.~Stuelpnagel, ``{On the Parametrization of the Three-Dimensional Rotation
  Group},'' \emph{SIAM Review}, vol.~6, no.~4, pp. 422--430, oct 1964.
  [Online]. Available: \url{http://epubs.siam.org/doi/10.1137/1006093}
\BIBentrySTDinterwordspacing

\bibitem{Choe2006}
S.~B. Choe, ``{Statistical Analysis of Orientation Trajectories via Quaternions
  with Applications to Human Motion},'' no. December, p. 117, 2006.

\bibitem{Goddard1997}
J.~S. {Goddard Jr.}, ``{Pose and Motion Estimation from Vision Using Dual
  Quaternion-based Extended Kalman Filtering},'' Ph.D. dissertation, 1997.

\bibitem{Goddard1998}
J.~S. Goddard and M.~A. Abidi, ``{Pose and Motion Estimation Using Dual
  quaternion-based Extended Kalman Filtering.}'' in \emph{Image (Rochester,
  N.Y.)}, R.~N. Ellson and J.~H. Nurre, Eds., vol. 3313, no. January, mar 1998,
  pp. 189--200.

\bibitem{Kavan2006}
L.~Kavan, S.~Collins, C.~O'Sullivan, and J.~Zara, ``{Dual quaternions for rigid
  transformation blending},'' \emph{Technical report TCDCS200646 Trinity
  College Dublin}, no. TCD-CS-2006-46, pp. 39--48, 2006.

\bibitem{Antone2001}
M.~E. Antone and S.~Teller, ``{Robust Camera Pose Recovery Using Stochastic
  Geometry},'' Ph.D. dissertation, 2001.

\bibitem{Love2007}
\BIBentryALTinterwordspacing
J.~J. Love, \emph{{Bingham Statistics}}.\hskip 1em plus 0.5em minus 0.4em\relax
  Dordrecht: Springer Netherlands, 2007, pp. 45--47. [Online]. Available:
  \url{http://dx.doi.org/10.1007/978-1-4020-4423-6{\_}19
  http://www.springerlink.com/index/10.1007/978-1-4020-4423-6{\_}19}
\BIBentrySTDinterwordspacing

\bibitem{Glover2011}
J.~Glover, G.~Bradski, R.~R. Rusu, and G.~Bradski, ``{Monte Carlo Pose
  Estimation with Quaternion Kernels and the Bingham Distribution},'' in
  \emph{Robotics: Science and Systems VII}.\hskip 1em plus 0.5em minus
  0.4em\relax Robotics: Science and Systems Foundation, jun 2011, p.~97.

\bibitem{Mardia2007}
\BIBentryALTinterwordspacing
K.~V. Mardia, C.~C. Taylor, and G.~K. Subramaniam, ``{Protein Bioinformatics
  and Mixtures of Bivariate von Mises Distributions for Angular Data},''
  \emph{Biometrics}, vol.~63, no.~2, pp. 505--512, jun 2007. [Online].
  Available: \url{http://doi.wiley.com/10.1111/j.1541-0420.2006.00682.x}
\BIBentrySTDinterwordspacing

\bibitem{Kraft2003}
E.~Kraft, ``{A quaternion-based unscented Kalman filter for orientation
  tracking},'' in \emph{Proceedings of the 6th International Conference on
  Information Fusion, FUSION 2003}, 2003.

\bibitem{Julier1997}
S.~J. Julier and J.~K. Uhlmann, ``{A New Extension of the Kalman Filter to
  Nonlinear Systems},'' in \emph{Proc. of AeroSense: The 11th Int. Symp. on
  Aerospace/Defense Sensing, Simulations and Controls}, vol. The 11th I, 1997,
  pp. 182--193.

\bibitem{Eidenberger2008}
\BIBentryALTinterwordspacing
R.~Eidenberger, T.~Grundmann, W.~Feiten, and R.~Zoellner, ``{Fast parametric
  viewpoint estimation for active object detection},'' in \emph{2008 IEEE
  International Conference on Multisensor Fusion and Integration for
  Intelligent Systems}.\hskip 1em plus 0.5em minus 0.4em\relax IEEE, aug 2008,
  pp. 309--314. [Online]. Available:
  \url{http://ieeexplore.ieee.org/document/4648083/}
\BIBentrySTDinterwordspacing

\bibitem{Bishop2007}
C.~M. Bishop, ``{Pattern Recognition and Machine Learning},'' \emph{Journal of
  Electronic Imaging}, vol.~16, no.~4, p. 049901, jan 2007.

\bibitem{Runnalls2007}
A.~R. Runnalls, ``{Kullback-Leibler approach to Gaussian mixture reduction},''
  \emph{IEEE Transactions on Aerospace and Electronic Systems}, 2007.

\bibitem{Brox2006}
\BIBentryALTinterwordspacing
T.~Brox, B.~Rosenhahn, U.~G. Kersting, and D.~Cremers, ``{Nonparametric Density
  Estimation for Human Pose Tracking},'' in \emph{Pattern Recognition,
  Proceedings}, 2006, pp. 546--555. [Online]. Available:
  \url{http://link.springer.com/10.1007/11861898{\_}55}
\BIBentrySTDinterwordspacing

\end{thebibliography}
\bibliographystyle{IEEEtran}

\end{document}